# A XGBoost Algorithm-based Fatigue Recognition Model Using Face Detection

Xinrui Chen[1], Bingquan Zhang[1,*]

[1]Department of Mathematics

Zhijing College of Zhejiang University of Technology.Shaoxin City,P.R.China

[*]zbq@zjc.zjut.edu.cn

**Abstract:**

As fatigue is normally revealed in the eyes and mouth of a person's face, this paper tried to construct a XGBoost Algorithm-Based fatigue recognition model using the two indicators, EAR (Eye Aspect Ratio) and MAR(Mouth Aspect Ratio). With an accuracy rate of 87.37% and sensitivity rate of 89.14%, the model was proved to be efficient and valid for further applications.

## 1 Introduction

People in a state of fatigue often experience symptoms such as low attention and sluggish response and negative consequences of various kinds due to fatigue are likely to occur. For example, fatigue driving has been found to be a chief cause for fatal traffic accidents. It is imperative to establish a fatigue recognition mechanism by face detection. Technically, the existing fatigue detection methods mainly use facial indicators such as eye movement [1] and yawn [2], and human body indicators such as heart rate [3] and blood pressure. This paper proposes to analyze facial state in the detection of fatigue. As face recognition technology has been rapidly developing in its maturity, it has been widely used in social practices. The technology of face recognition with a camera has made it possible to detect fatigue by facial indicators.

## 2 Fatigue detection by facial state indicators

Since many features of the face can more sensitively display conditions of human fatigue than features in the body, the recognition of fatigue by facial features has greater reliability. Methodologically, this paper used the dlib detector and predictor in the Python system to detect the specific location of facial features, and employed the OpenCV package to extract the data.

## 2.1 Eye Aspect Ratio(EAR) for Eye Detection

Eye Aspect Ratio(EAR) is a measurement commonly used for fatigue recognition by eye state. It refers to the ratio of two distances, namely, the vertical distance between the points on the upper and the low sides of the eye and the horizontal distance between the two points at the left and the right sides of the eye. Its calculation is given in the following formula:

$$EAR = \frac{\|P_2 - P_6\| + \|P_3 - P_5\|}{2\|P_1 - P_4\|}$$

In the formula, the meanings(value) of each variable are shown in the figure below, where p2 and p6, p3 and p5 form two pairs of features locating the vertical distance while p1 and p4 form a pair locating the horizontal distance.

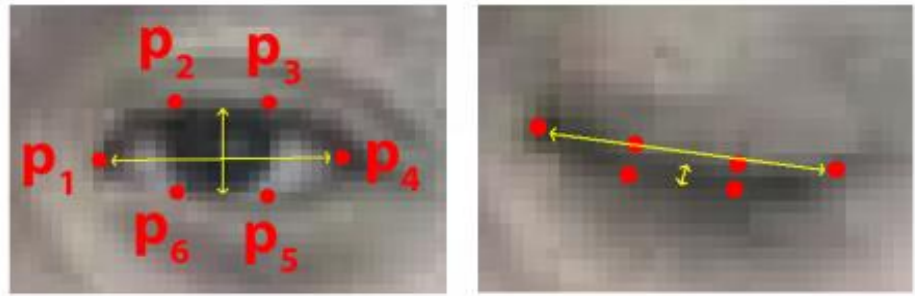

Fig 2.1 Locations of 6 key points in the eye

When the eyes are open, the value of EAR is almost unchanged, but when the eyes are closed, the EAR is almost 0. Therefore, in the experiment where static pictures were taken, we could judge whether the subject was squinting or even dozing off due to fatigue through the EAR value; in the video stream, we could set a threshold for EAR (such as 0.75 ), and use this as an indicator to calculate the blink frequency to get the fatigue state of the eyes.

## 2.2 Mouth Aspect Ratio (MAR) for mouth detection

In addition to the detection of fatigue by eye movements, yawning or opening the mouth are usually positively correlated with the fatigue state of subjects. Mouth Aspect Ratio (MAR) is also an index commonly used to detect fatigue. Like EAR, it uses the values of two distances, the height and the length between the identified points around the mouth, and with the ratio of the two distances, it helps determine the degree of opening and closing of the mouth.

$$MAR = \frac{\|P_2 - P_6\| + \|P_3 - P_5\|}{2\|P_1 - P_4\|}$$

In the above formula, the meanings (value) of each variable (point) are shown in the figure below. In the experiment, we measured the inner contour of the lips so as to eliminate the influence of the thickness of subjects' lips in the MAR calculation and try to ensure accuracy of measurement to a certain extent.

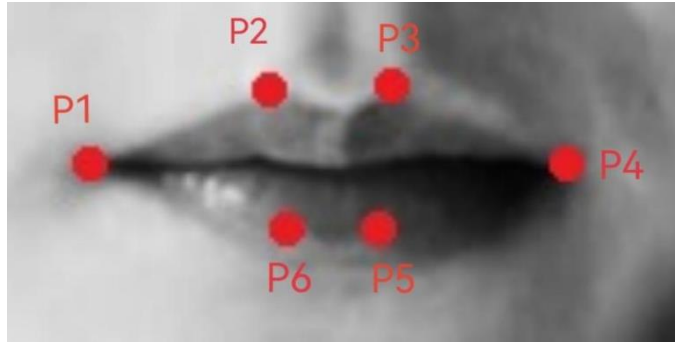

Fig 2.2 Location of key points in the mouth state

Similar to the case of EAR, in a given frame of a video stream or any static picture, we can set a threshold (such as 0.5) for MAR to independently determine whether there are yawning and other behaviors in the mouth state.

## 2.3 A fatigue detection model based on XGBoost

The preceding sections introduce methods of detecting fatigue in a certain part of the face, but fatigue detection made in integration through multiple facial indicators has more important practical significance. Therefore, we build a model based on two indicators of face fatigue. For the collected data of the facial state presented in the preceding sections, we used the XGBoost model to compute them. XGBoost, which is a kind of Gradient Boosting Decision Tree (GBDT), has many advantages to offer comparatively. First, the accuracy of prediction is higher, and the sensitivity to noise and multicollinearity in the data is lower; second, it has more advantages in mining variables, and its unique tree structure enables researchers to obtain hidden variables from original variables through combination [5]. So it is suitable for abstract data modeling such as the case for multiple indicators of the face in the study.

Being an optimized distributed gradient boosting, it has been widely acclaimed for its efficiency, flexibility and portability. It is an improvement of the gradient boosting algorithm [6]. Different from Newton's method which is used to solve the extreme value of the loss function, XGBoost not only expands the loss function Taylor to the second order, but also adds a regularization term to it. The objective function during training consists of two parts, the first part of which is the gradient boosting algorithm loss, and the second of which is the regularization term. Its loss function is:

$$L(\emptyset) = \sum_{i=1}^{n} l(y_i', y_i) + \sum_{k} \Omega(f_k)$$

In the above formula, $n$ is the number of samples, $l$ is the loss of a single sample; assuming it is a convex function, $y_i'$ is the predicted value of the model for the samples, and it is the real value of the sample under study.

Next, to calculate the derivative of the loss function for the predicted value of each sample, the XGBoost model performs a second-order Taylor expansion on the loss function:

$$\hat{y}_i^{(0)} = 0; \hat{y}_i^{(1)} = f^1(x_i) = \hat{y}_i^{(0)} + f^1(x_i)\ldots;$$

$$\hat{y}^{(t)} = \sum_{k-1}^{t} f^k(x_i) = \hat{y}^{(t-1)} + f^t(x_i)$$

In the above formula, CART model represents the kth tree.

In order to prevent the increased risk of overfitting, a penalty term is added to the objective function:

$$\Omega(f^k) = \gamma T + \frac{1}{2}\lambda \sum_{j=1}^{T} \omega_j^2 f$$

In the above formula, $\lambda$ Indicates the degree of punishment, T the number of leaves, and $\omega$ the weight of leaf nodes.

Thus, we can finally get the complete objective function as follows:

$$\text{obj}^{(i)} = \sum_{k-1}^{n} L(y_i, \hat{y}_i^{(t-1)} + f^t(x_i) + \Omega(f^k))$$

The model was implemented in an experiment with the Python program. First of all, in the face detection, a total of 68 face feature points were located and identified using the dlib detector. The images of the face state were made gray so as to reduce the interference of color in the analysis. Second, as regards the identified eye and mouth positions, calculated the EAR and MAR values and entered them as the input for variables of the XGBoost model, and took fatigue state as the output variable. The decision tree of the model was 2000, the depth was 6, and binary:logistic was used as the objective function. 70% of the data was selected as the training set and 30% of the data as the test set. The prediction results obtained are as follows in Table 2.1.

Table 2.1 Results of the Detection Model

| Actual Fatigue State | Predicted Fatigue State | Predicted Non-fatigue State |
|---|---|---|
| Yes | 1626 | 198 |
| No | 304 | 1850 |

Accuracy is a most concerned attribute for the validity of a model.

$$Accuracy = \frac{TN + TP}{TN + FN + TP + FP}$$

In addition, since it is more important to detect fully the state of fatigue in the face fatigue detection model, sensitivity is selected here to verify the model. The formula of sensitivity calculation is as follows:

$$Sensitivity = \frac{TP}{TP + FN}$$

The calculated accuracy is 87.37%, and the sensitivity is 89.14%. It can be seen therefore that both the accuracy and sensitivity of the model are relatively high, and the fatigue state of the subjects can be predicted relatively accurately based on the face data of the static picture. It can be concluded accordingly that it would have better performance in fatigue discrimination in life.

## 3 Results and Discussion

According to the experimental results, the XGBoost model can be used to determine with efficiency EAR and MAR respectively. Therefore, given any static picture of the face state, the proposed model can be used as a reference for detecting fatigue status by eye and mouth data.

In addition, although the use of XGBoost algorithmon in human facial fatigue detection in the study is based on static images, the model can be surely extended to analyze video streaming data. In this regard, first of all, the video stream may be decomposed into frames by using certain technologies; secondly, the XGBoost algorithm can be used to judge the fatigue state of each frame in a certain video, and then the probability of each frame being determined as fatigue can be counted, and the mean number of the probability value can be calculated and finally the face fatigue status in a certain video stream could be obtained.

To conclude, the method proposed in this paper provides a relatively accurate model for detecting facial fatigue, which can help solve the problem of detecting fatigue in real life, and may have the potential to be put into practical applications.